# Learning From the Pros: Extracting Professional Goalkeeper Technique from Broadcast Footage

Matthew Wear[1], Ryan Beal[1,2], Tim Matthews[2], Tim J. Norman[1], and Sarvapali D. Ramchurn[1,2]

[1]University of Southampton, Southampton, UK
[2]Sentient Sports, UK

## 1. Introduction

Over the past few years there has been an increase in research into improving sports analytics using artificial intelligence (AI)-based techniques [1]. Much of this research has focused on extracting contributions of outfield players [2,3] or around team tactics [4,5]. However, there are fewer examples of research focused on the analysis of player technique, and in particular the technique of goalkeepers in soccer. This is especially the case when it comes to looking at lower-level players in grassroots sports where there are millions of players across the world rather than just creating models for the top 1% in the professional game.

Of the millions of goalkeepers worldwide, only 20 plays regularly in the English Premier League (EPL). As an amateur goalkeeper playing grassroots soccer, who better to learn from than top professional goalkeepers? Advances in computer vision and pose estimation using deep learning models has introduced an opportunity to learn about soccer directly through video footage. Therefore, in this paper we present a new model to use broadcast footage which allows us to analyse goalkeeper techniques in two key situations they will face: penalties and one-on-ones (1v1s). In doing so, we extract value from professional techniques for amateur players and coaches to learn from. Our model also provides them with an open-source framework to evaluate their own techniques using basic equipment.

Building on previous black-box analyses [6], we use 3D body pose data from broadcast footage, as well as event data to learn professional save technique. In order to discover goalkeeper save technique, we employ unsupervised machine learning algorithms to group together similar saves. Output from the unsupervised learning model is then used to train a white-box "*expected saves*" (xS) model, from which we can identify the optimal goalkeeper technique in different match contexts. In summary, we make the following novel contributions:

- A new open-source dataset of saves and body pose from broadcast footage[1]
- An analysis of techniques employed by professionals in different match contexts
- A white-box xS model that can be used to derive teachable insights for professional and grassroots players alike

The work presented in this paper has several practical applications. Firstly, the methods can be used to produce optimal technique maps by comparing the value of employing particular save techniques. These optimal technique maps can be used as training material to help goalkeepers and coaches alike. Goalkeepers will be able to see general rules for when to use different save techniques and use these to improve their game. In Section 6.4, we show how our models can be employed as a scouting tool by evaluating Premier League goalkeepers on their ability to frequently employ optimal save technique. In Section 6.6, we will show examples of how grass-roots goalkeepers can utilise our models using self-recorded amateur footage to compare their technique choice to Premier League and International goalkeepers.

---

[1] https://github.com/MattWear21/LearningFromThePros.

## 2. Background

Power et al. [6] used 2D pose estimation to analyse the pose attributes that best predict where the striker will place the ball from a penalty kick. It was shown, for example, that when the striker's hip angle is open towards the right-hand post it indicates a larger probability of shooting to the right. Pose estimation has also been applied in basketball to analyse the shooting style of basketball players in three-point shots [7]. They discovered that Stephen Curry takes a higher proportion of off-balance shots compared to the average player. Pose estimation techniques can not only be used for technique analysis but has also found use in orientation prediction in soccer [8,9].

Previous work in goalkeeper analytics has mostly focused on metrics for evaluating professional goalkeeper's ability in different facets of their game. Work presented in [10] introduces a goalkeeper evaluation framework that uses event data to value actions. The paper introduces methods that provide metrics for shot stopping ability, positioning, cross collection, and distribution. For evaluating the effectiveness of goalkeepers in saving shots they introduce a variant of the xG model [11] called Post-shot Expected Goals (PSxG).

Work in this paper utilises expert-defined save techniques when analysing goalkeeping technique in 1v1 situations.[2] We define two save techniques that are used when the goalkeeper is in a "*ready pose*" position. These are the "*passive set*" and "*aggressive set*". The only difference in the two being that in the passive set the goalkeeper is opting to stay close to their goal line. On the other hand, the aggressive set is where the goalkeeper is in a ready pose but is engaging on the striker. The "*spread*" technique is where the goalkeeper is engaging on the striker and making their body as wide as possible to cover as much of the goal as they can. Finally, the "*smother*" technique is where the goalkeeper gets low to the ground and tries to save the ball at the striker's feet.

## 3. Data

The models introduced in this paper are trained and evaluated on image and event data from soccer matches. We use broadcast footage from 764 matches in the 2018 World Cup and 2 EPL seasons (2017-2019) to assemble a dataset of 1v1 and penalty saves. A total of 590 images of goalkeepers reacting to 323 unique 1v1 shots were collected, as well as one image in each of the 369 penalty shots available for collection. Each image corresponds to the frame in the broadcast footage (both replays and live footage) at which the striker first makes contact with the ball when shooting, as this is the instance when the goalkeeper should be in their ready pose [12]. The goalkeeper's entire body was included in each image, and where body parts are occluded, the area in which these occluded parts were likely to be are included in the image. Figure 1 shows examples from the goalkeeper image dataset.

---

[2] 1v1 Save Technique definitions provided by John Harrison (https://twitter.com/Jhdharrison1).

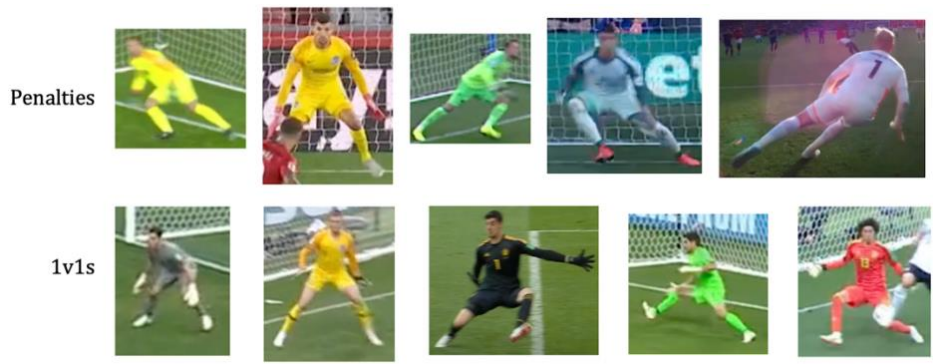

**Figure 1**: Sample of images that form our dataset of penalties and 1v1s.

We use features derived from StatsBomb[3] event data in combination with learned save technique predictions to train our xS model in 1v1 situations. The event data gives a record of every action that occurred during a soccer match, as well as the pitch coordinates of the ball and players at the time each shot was taken. In this paper, we define a 1v1 as a shot for which only the goalkeeper is inside the triangle formed by joining the shot location, the right post, and the left post. Also, the striker must be higher up the pitch than any other defending player apart from the opposition goalkeeper and the shot is made with the player's foot. Three examples of 1v1s are shown in Figure 2.

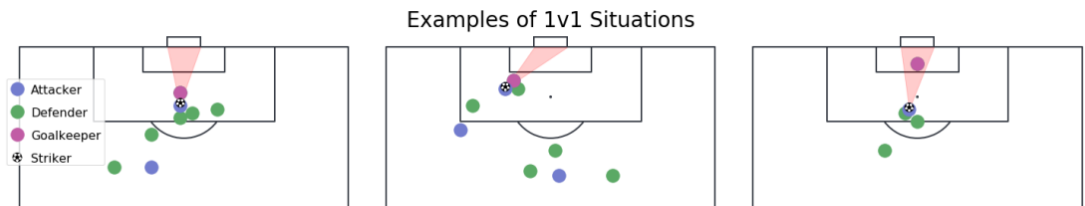

**Figure 2**: Examples of 1v1s according to our definition. Only the goalkeeper can be in the triangle formed between the striker and two goal posts.

# 4. 3D Pose Estimation

In this section, we describe how we extract the 3D pose of goalkeepers when they make a save. We discuss how these can be normalized for the camera angle in Section 4.2 and introduce the goalkeeper engagement metric (GKEM) in Section 4.3, the metric designed to quantify the extent to which a goalkeeper is closing down the striker.

### 4.1. Estimation Model
The 3D pose estimation model (PoseHG3D) introduced in [13] is utilised to extract 3D body pose coordinates from the goalkeeper images. PoseHG3D is a convolutional neural network that is pre-trained on a dataset of "in-the-wild" images of humans, and its suitability for the task of goalkeeper images is evaluated in this paper. 3D pose estimation allows for the normalisation of each body pose to approximate a single effective camera angle (full method described in Section 4.2). There were a variety of options for a 3D pose estimation model [14,15,16,17], but PoseHG3D suited the task of prediction on goalkeeper images for the following reasons:

- **Body Parts:** Prediction of the location of all body parts, including those that are occluded by other players or objects.
- **Scale:** Due to its use of geometric constraints on bone lengths in the human skeleton, all predictions are made on the same scale. For example, a crouched goalkeeper should have a smaller height in the resulting coordinate system than a goalkeeper standing upright.
- **Real-World Results:** The model is proven to achieve good results on in-the-wild images.

---

[3] https://statsbomb.com.

Given an input image with the goalkeeper centered and the main object, the model predicts both the 2D pose in image coordinates, as well as the 3D pose. The body pose is represented by 16 body pose keypoints (see Appendix 1). The origin of each body pose is then chosen to be located at the body pose keypoint representing the middle of the goalkeeper's hip. Figure 3 shows an illustration of the extraction of 2D and 3D body pose coordinates on Ederson Moraes saving a penalty.

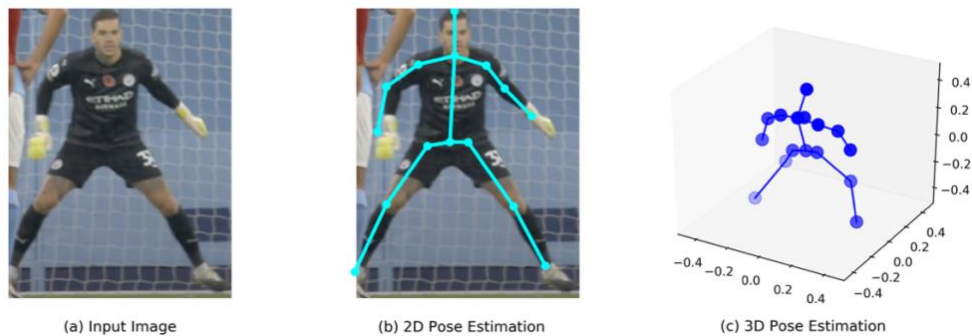

**Figure 3**: The 3D pose estimation model predicts both the 2D and 3D pose coordinates.

Body pose predictions were made on all images of penalty saves and 1v1 saves. A total of 181 of the 590 1v1 images, and 136 of the 369 penalty body poses were removed due to them being subjectively deemed to be poor representations of the true body pose. Typically, the cause of bad results from the pose estimator were due to low-resolution images or significant body part occlusion. Figure 4 shows six examples of the final body-poses that we extracted using the PoseHG3D model.

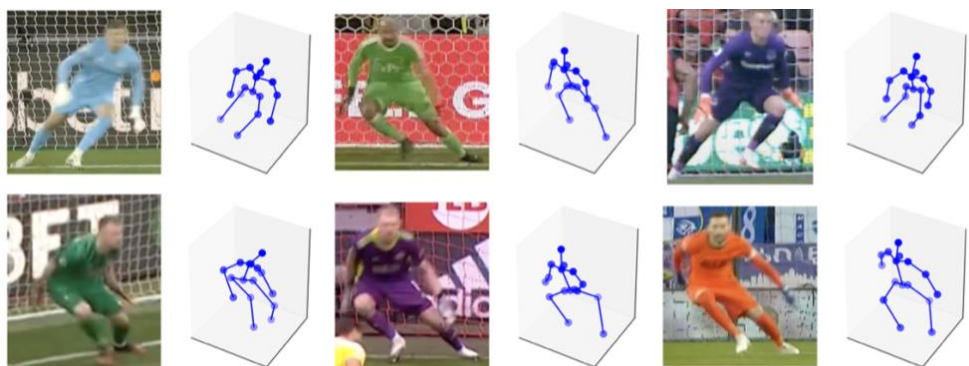

**Figure 4**: Examples of the 3D pose predictions from PoseHG3D.

## 4.2. View-Invariance

The camera angles used to capture saves by goalkeepers vary from stadium to stadium, and there are often multiple cameras covering a soccer match. Therefore, the body pose coordinates depend on the angle from which the image was captured. To allow for direct comparison of body poses, we create a view-invariant dataset of pose coordinates.

A goalkeeper will attempt to maximise their coverage of the goal [12], and we utilise this in order to create a view-invariant body pose. Our method of deriving the view-invariant body pose is to rotate each pose about the y-axis (vertical axis) such that the goalkeeper's width is maximised. In doing so, the effective camera angle for every body pose approximates that of the striker's point of view looking directly at the centre of the goal. A rotation of θ degrees is achieved by the matrix multiplication of the rotation matrix, $R_y$, and each of the 16 body pose coordinates, $p_j$. The rotation matrix is defined in equation (1).

$$R_y = \begin{pmatrix} \cos\theta & 0 & \sin\theta \\ 0 & 1 & 0 \\ -\sin\theta & 0 & \cos\theta \end{pmatrix} \qquad (1)$$

The rotated body pose coordinates are then given by $P_\theta$, shown in equation (2).

$$P_\theta = \begin{pmatrix} (R_y \mathbf{p}_1^T)^T \\ (R_y \mathbf{p}_2^T)^T \\ \vdots \\ (R_y \mathbf{p}_{J=16}^T)^T \end{pmatrix} \qquad (2)$$

For each body pose, we calculate $P_0, P_{10}, P_{20}, \ldots, P_{90}, P_{270}, P_{280}, \ldots, P_{350}$, and keep the one for which the goalkeeper's width is maximised on the 3D to 2D projection onto the x-y plane. $P_\theta$ for which the goalkeeper width is maximised we shall denote as $P_A$.

There are several instances of images taken from behind the goalkeeper in our dataset and to accommodate these images we must rotate their corresponding body poses by 180 degrees to be in line with all other images. All body poses, $P_A$, for which the right hand x-coordinate is greater than the left hand x-coordinate is deemed to be a photo taken from behind the goalkeeper. Figure 5 gives an example of the view-invariance method in action.

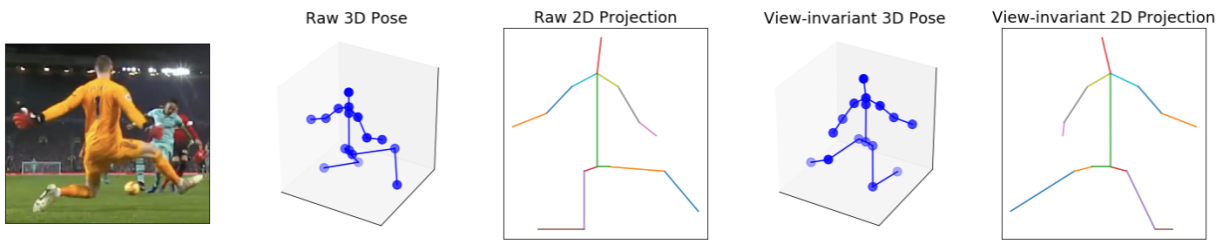

**Figure 5**: Illustration of the view-invariance method, shown using both the 3D pose and 3D to 2D projection.

## 4.3. Goalkeeper Engagement Metric

Through the use of 3D body pose data and view-invariance, we have a good representation for the technique that a goalkeeper used to make a save. However, in the case of 1v1s, a key decision that a goalkeeper must make is whether to close down the striker. Some goalkeepers such as Ederson and Manuel Neuer are known for their style of coming far off their line to make 1v1 saves, but others prefer to stay closer to their goal line. The decision of when and how far to come off the goal line is a key skill for amateur goalkeepers to learn.

We propose a goalkeeper engagement metric (GKEM) that describes the extent to which the goalkeeper closes down the striker as they shoot the ball. Derived using event data, GKEM is the ratio between the distance from striker to goalkeeper and striker to the centre of the goal. Let the goalkeeper's position in pitch coordinates be $(x_{gk}, y_{gk})$, the striker's position be $(x_s, y_s)$ and the centre of the goal to be $(x_{cg}, y_{cg})$. Then, GKEM is given by equation (3).

$$\text{GKEM} = \frac{\sqrt{(x_s - x_{gk})^2 + (y_s - y_{gk})^2}}{\sqrt{(x_s - x_{cg})^2 + (y_s - y_{cg})^2}} \qquad (3)$$

A large GKEM indicates when the goalkeeper is not engaging on the striker and is in a passive position. Small GKEM values indicate where the goalkeeper is engaging on the striker. Some examples of GKEM in 1v1s are shown in Figure 6.

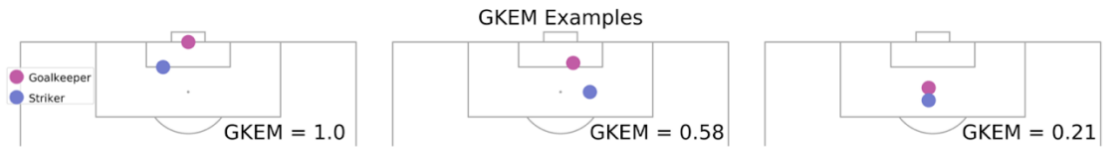

**Figure 6**: Examples of the GKEM metric in three different 1v1 situations.

The GKEM corresponding to every 1v1 in our dataset is calculated and will subsequently be used with body pose features to learn a grouping of save technique in Section 5.1. GKEM is not calculated for penalties because the location of the goalkeeper does not exist in event data for penalties and the variation in goalkeeper location will be minimal because they are required to be on the goal line when the penalty is taken.

# 5. Learning Save Technique

In this section, we investigate whether we can learn goalkeeper save technique from the professionals using only body pose and event data. We seek to learn a grouping of body pose and GKEM where each group corresponds to a particular type of save technique. To achieve this, we will use K-Means clustering [18], an unsupervised machine learning algorithm. Unsupervised learning lends itself to this task of grouping saves as it does not require labelled data, and instead learns directly from the features.

## 5.1. 1v1 Saves

For 1v1 saves we use a dataset that consists of the 3D body pose coordinates. However, when grouping together 1v1 save techniques, there is little informative value in body pose depth, unlike in penalties. Variation in goalkeeper poses in 1v1 situations is large enough such that little information in depth is added from what can be learned from a 2D pose. For this reason, we decided to remove the depth (z-coordinate) of each body pose and instead use the 2D projection onto the x-y plane instead. This reduces the dimensionality of the problem. Each body pose is then described by a vector of length 33 (2D body pose + GKEM).

The K-Means algorithm is trained using K-Means++ initialisation [19]. Four cluster centres were found to be optimal in identifying granular differences in save technique based on subjective analysis of saves in each cluster aligning with domain knowledge. The clustering algorithm is applied to a 33-dimensional dataset, so to visualise the results from K-Means, we need to use a dimensionality reduction technique. We apply t-distributed stochastic neighbor embedding (t-SNE) [20] to reduce the data to two dimensions for visualisation purposes. Each point in Figure 7 represents a save and is colored by their save technique found using K-Means.

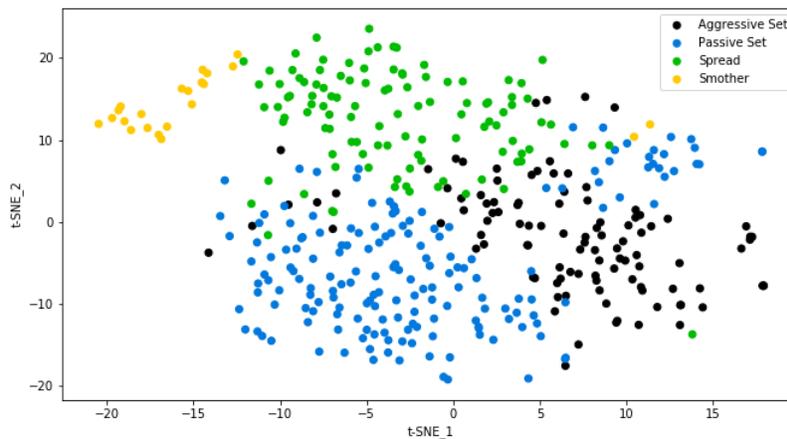

**Figure 7**: Each point is a 1v1 save represented in 2 dimensions, using t-SNE. The saves are colored by their save technique as derived from the K-Means clustering algorithm.

To get a better understanding of how K-Means has separated the saves and what save techniques it found, we extract the save that best represents each of the four cluster centres that K-Means converged to. Figure 8 shows the saves that are closest in Euclidean distance to their respective cluster centres.

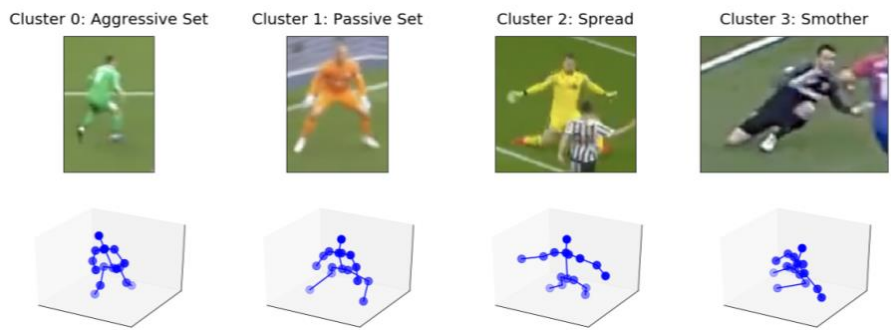

**Figure 8**: The saves shown are those that are closest to each K-Means cluster center by Euclidean distance.

Cluster 1 contains goalkeepers using the passive set position, which is the most common save technique, being used in 172 of the 409 1v1s analysed. It is most employed by a goalkeeper in long-distance 1v1s or when there is sufficient cover from defenders. Cluster 0 holds the aggressive set; a very similar body pose to those in cluster 1 but where the goalkeeper is further engaged with the striker. The GKEM corresponding to clusters 0 and 1 is 0.62 and 0.78, respectively. Cluster 2 holds the spread technique, where the goalkeeper does not have enough time to react to the trajectory of the shot so instead tries to cover as much of the goal as possible. Cluster 3 is the smother technique, which is the rarest type of technique used in 1v1 situations, being used in just 5% of 1v1s in our dataset. Here, the goalkeeper gets into a very low position and is often used when dealing with shots coming from a very tight angle to goal.

For amateur goalkeepers, the findings in this section of the paper can be used as an understanding of the typical techniques used by professional goalkeepers. Learning these techniques can be valuable for upcoming goalkeepers and they can also look at footage of their own game to see if they ever stray from these four techniques used by professionals. Where the real value lies for amateur goalkeepers is understanding when to use which technique and this is what is investigated in Sections 6.2 and 6.3.

**5.2. Penalty Saves**
Samir Handanovic has a penalty record of 42 saves in 122 penalties[4], a total contrast to David de Gea who failed to save 40 consecutive penalties. In the 2014 World Cup Louis Van Gaal decided to substitute Jasper Cillessen for Tim Krul, solely for his ability to save penalties. Clearly, there is significant value in amateur goalkeepers learning penalty saving techniques from professionals as it is a crucial part of being a complete goalkeeper.

Just as for 1v1s in Section 5.1, we will use K-Means to identify penalty saving techniques used by professional goalkeepers. An issue we face when clustering penalty saves by the goalkeeper's body pose is that unsupervised methods simply group them into left and right dives, providing little insight of value. Instead, we transform the 3D body pose into a hand-crafted 5-dimensional feature space. This feature space is designed to be invariant to the direction in which the goalkeeper chose to dive. This way we force the model to learn differences in technique rather than direction choice. The chosen hand-crafted features are torso angle, body angle, body height, forward step distance, and hand height (see Appendix 2 for precise definitions).

On application of the K-Means algorithm with K-means++ initialisation [19], two clusters were found to be optimal for clustering the penalty save feature space, as two clusters achieved the largest silhouette score [21]. Figure 9 shows examples of saves found in each of the two clusters. Cluster 0 (consisting of 86 instances) has dives where the goalkeeper is significantly into the diving action when the ball is struck. Cluster 1 (consisting of 147 instances) corresponds to dives that are either upright or slightly leaning when the ball is struck. The key difference in features between the two clusters is that the average torso and body angle for cluster 0 is 36.5 and 27.5 degrees, compared to 10.8 and 16.4 degrees for cluster 1. Digging deeper, we can see that the average save percentage for cluster 0 and cluster 1 is 18.6% and 15.0%, respectively. This provides some evidence, albeit from a relatively small sample size, that increasing your body angle at the point of impact of the strike increases the chance of saving a penalty.

---

[4] https://www.transfermarkt.co.uk/samir-handanovic/profil/spieler/28021.

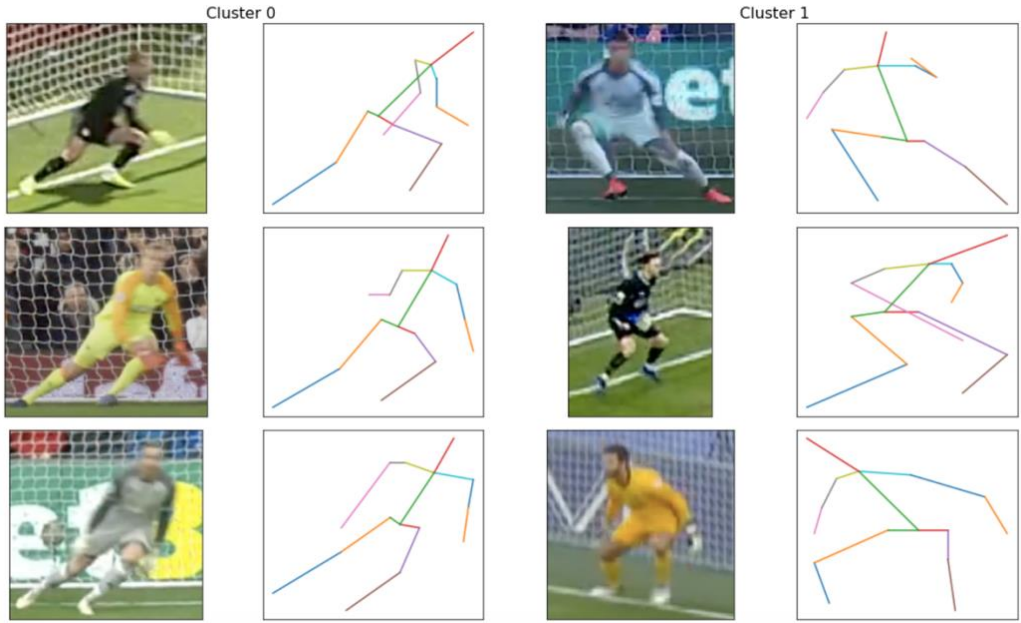

**Figure 9**: Examples of penalty saves that belong to the two clusters discovered using K-Means.

# 6. Applications

In this section, we describe several real-world applications for the models that are presented in this paper. We discuss how our models can add value to goalkeepers both in the professional game and for amateurs.

### 6.1. Post-Match Save Analysis

In Section 5.1, we applied unsupervised learning to identify four distinct save techniques used by professional goalkeepers to save 1v1 shots. In this section, we quantify the effectiveness of each technique depending on the type of 1v1 shot being faced. Once we can assign a value to each save technique, we can then build save technique maps that describe which technique is optimal depending on the distance and angle of the shot, as well as match context. Amateur goalkeepers and coaches can then use these optimal technique maps to help improve their own game.

Here, we use a supervised machine learning approach to build an "*expected save*" (xS) model. Specifically, we train a "Support Vector Machine" (SVM) model [22] which is a classification algorithm that we have selected as it is able to achieve good generalisation capacity from relatively few training instances. Platt scaling [23] was used to derive probabilities from the predictions of the SVM. To effectively evaluate save technique we filter the 1v1s to those that were on target (either a goal or saved by the goalkeeper). The target variable is a binary indicator showing whether the shot was saved or scored. To provide context of the shot, we derived the distance and angle to the goal, as well as whether the striker was under pressure from a defender or not, all of which act as input features in our xS model. The K-Means algorithm described in Section 5.1 outputs a cluster membership label for each save, which is also added as a categorical feature to describe the save technique used in each 1v1.

The dataset of 311 on-target 1v1s is separated into training and test sets using a 70/30 split. A grid search with 5-fold cross-validation was performed on the training set to find the optimal hyperparameters for the model[5]. The model was then refitted on the full training set using these optimal hyperparameters and achieves an accuracy of 68.5% on the test set. This result shows that the SVM model is more predictive than a random guess as to the outcome (goal/save) of a 1v1 shot, since 47.6% of on-target 1v1s in the training set were saved.

Figure 10 shows Neil Etheridge closing down Jamie Vardy using the smother technique. In doing so he has a 47.4% chance of saving the shot according to our xS model. This is the worst option of the four save techniques, according to our model. A passive set would have seen him have an 11% higher chance of making a save. Figure 10 also shows Bernd Leno in too high a position against the opposing striker. He is using the aggressive set technique as he is in a ready pose position, but his GKEM is low at 0.56. The model predicts a 38.7% chance that he makes the save in this situation, assuming the shot is on target. His chances would have increased significantly to 46.2% had he stayed closer to his line, adopting a passive set position.

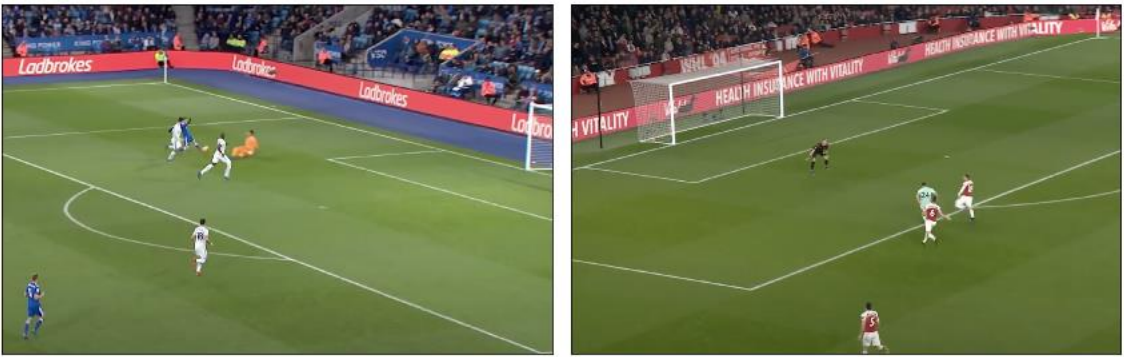

**Figure 10**: Neil Etheridge saving a 1v1 using the smother technique (left). Bernd Leno facing a 1v1 using the aggressive set technique (right).

## 6.2. Appraising Save Technique

To evaluate the effectiveness of each save technique for a given 1v1 situation, we can compare their xS probability. To aid with this comparison, we introduce the expected saves above average (xSAA) metric which describes how much more likely a goalkeeper is to make a save when employing that technique than the average xS for that shot. Specifically, we calculate the xS of a given shot for each of the four save techniques. The xSAA for a particular save technique is then its xS minus the average of all four xS values. Therefore, a positive xSAA indicates a save technique that will result in more saves than expected, and a negative xSAA indicates a save technique that will result in fewer saves than expected.

Figure 11 shows the xSAA map for the xS model. The area of the pitch in which the striker was when he took the shot is colored by its xSAA according to our model. Each pitch shows the xSAA for when the goalkeeper used a specific save technique and whether the striker was under pressure or not. The passive set significantly increases the save probability when the shot is outside the 6-yard box and can achieve an xSAA of up to 0.12. The spread technique is most effective against close, central 1v1s, but ineffective when the shot is from a very tight angle. From these tight angles using a spread technique can see a xSAA of -0.1. The spread technique is 10% more likely to result in a save than the passive set when the shot occurs in a central position, 4 yards out from goal. From tighter angles and close shots, an aggressive set achieves an xSAA of approximately 0.17. The smother technique is generally not preferred anywhere on the pitch but is most effective against shots outside the 6-yard box, or from a very tight angle to goal.

---

[5] Optimal hyperparameter set for the SVM model was an RBF kernel and regularisation parameter, C=1.

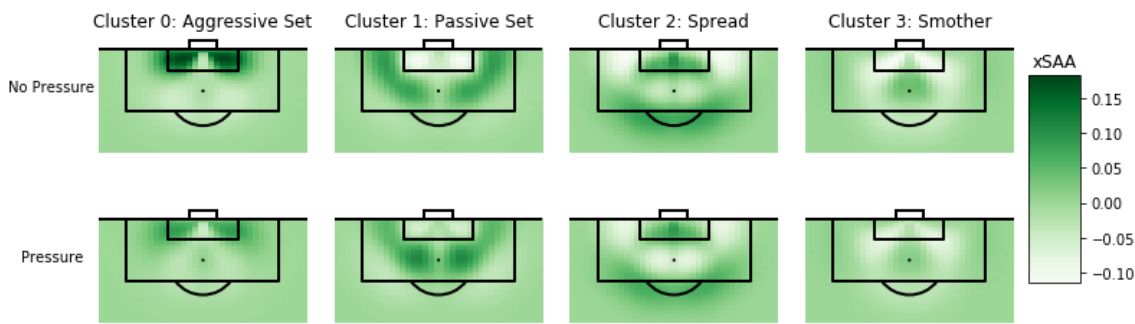

**Figure 11**: The pitch is colored by the xSAA for employing that specific save technique when the striker is taking a 1v1 from that position. 1v1 situations are divided by whether the striker was under pressure from a defender or not.

When the striker is under pressure, changes occur to the two set techniques. The passive set technique is still very effective against 1v1s outside the 6-yard box but is now more effective for close-range shots too than it was when the striker is not under pressure. The aggressive set technique is now less effective in all situations compared to when the striker is under no pressure but is still effective at close distances and tight angles to goal.

## 6.3. Extracting Optimal Save Technique

By selecting the save technique which maximises xS for a given location for the striker on the pitch, we can produce optimal technique maps. These plots show the technique the goalkeeper should use given the position of the striker and the pressure the striker is on. These can be provided as a coaching tool for amateur goalkeepers by giving them a good rule of thumb as to when to use the different techniques found in Section 5.1. Figure 12 shows the optimal technique maps for when the striker is under pressure and when they are not under pressure.

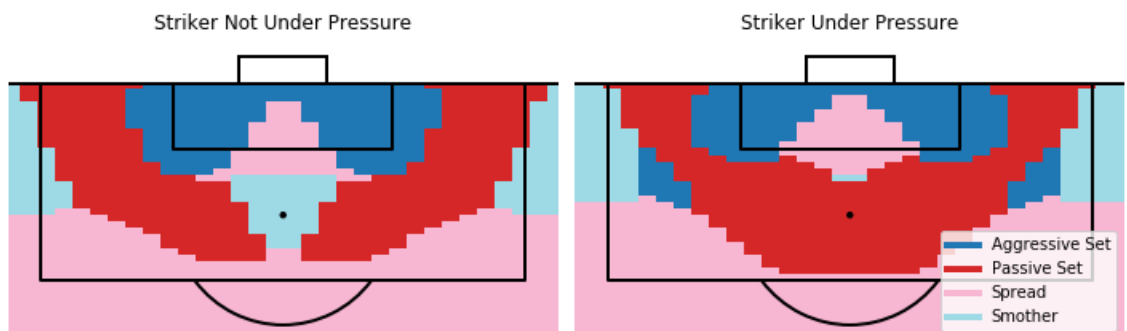

**Figure 12**: Optimal technique maps derived from our xS model. Areas of the pitch are colored by the technique a goalkeeper should employ when facing a 1v1 from that position on the pitch.

Some interesting discoveries are found in the optimal technique maps. Firstly, the passive set is optimal for long-distance 1v1s (further than 8 yards from goal) and should be used for shots as close as 7 yards if the striker is under pressure and not central to goal. In close central areas to the goal, the spread technique is the most effective. At these close distances, a goalkeeper will have minimal or no chance to react to the trajectory of the shot, so they should make themselves as wide as possible. The aggressive set is only optimal for close-range 1v1s at narrow angles to the goal. However, the aggressive set should be used more sparingly when the striker is under pressure from defenders. The model is unable to generalise on the edge of the penalty area due to a lack of data points in these areas of the pitch, so findings from these areas should be ignored. We know from domain knowledge that the light blue and pink sections outside the penalty area should in fact be a passive set.

### 6.4. Goalkeeper Evaluation

Current state-of-the-art in goalkeeper scouting using data and statistical methods are described in the goalkeeper evaluation framework presented in [10]. A PSxG model is used to measure a goalkeeper's ability to make saves, but so far there is no quantitative method to evaluate the ability for goalkeepers to choose optimal technique.

In this section, we use the xS model to rank professional goalkeepers by their ability to utilise the optimal save technique in 1v1 situations, thus adding another metric to a scout's toolkit. The metric being used to rank goalkeepers is the percentage of 1v1 situations in which the goalkeeper uses the optimal save technique. Table 1 shows this metric for each goalkeeper to have faced at least 15 1v1s in the Premier League 2018/19 or World Cup 2018. It should be noted that some goalkeepers often use non-optimal techniques but still get good results using their own style. For example, Peter Schmeichel's famous starfish technique[6] would have been unseen by our model, but he found the technique very successful.

**Table 1**: Professional goalkeepers in the EPL ranked by their ability to use the optimal save technique, according to our model.

| Goalkeeper | Optimal Save Technique (%) | 1v1s Faced |
| --- | --- | --- |
| Neil Etheridge | 62.5 | 16 |
| Lukasz Fabianski | 54.6 | 28 |
| Ben Foster | 40.0 | 20 |
| Kasper Schmeichel | 40.0 | 20 |
| David de Gea | 27.8 | 18 |
| Hugo Lloris | 20.0 | 15 |
| Sergio Rico | 15.0 | 20 |
| Jordan Pickford | 11.1 | 18 |

### 6.5. Identifying Significant Body Pose Features in Penalties

Analysis in [6] uses body pose data to understand the features that are significant predictors in where the striker will place the penalty. They found that when the striker's hip angle is open towards the right-hand post it indicates a larger probability of shooting to the right. In this paper, we extend this research by investigating whether any body pose features are significant predictors of penalty save probability. We use all the features in the penalty feature space to train a logistic regression model. The model is designed to predict the probability of a penalty save being made given the features in the feature space.

---

[6] https://www.90min.com/posts/peter-schmeichel-the-great-dane-who-re-invented-modern-goalkeeping.

A total of 225 of the 233 penalties in the dataset were on target, so 8 penalties were removed. The dataset was randomly split into a training set and testing set. A 70/30 split was used leaving 157 data points in the training set and 68 data points in the test set. The model achieves an 88.2% accuracy on the test set, but this is simply due to it always predicting a goal to be scored. However, variation in the predicted save probability exists, with this being as high as 32.9% and as low as 10.5%, with an average save probability of 17.7%. For comparison, 16.3% of saves in our dataset were saved. Table 2 shows the regression summary.

**Table 2**: Logistic regression summary when classifying penalty saves using the penalty save feature space.

| Feature | Coefficient | Std. Error | z value | Pr(|z|) |
|---|---|---|---|---|
| (Intercept) | -1.4829 | 0.210 | -7.073 | 0 |
| Torso Angle | 0.0447 | 0.263 | 0.170 | 0.865 |
| Body Height | -0.1887 | 0.249 | -0.759 | 0.448 |
| Forward Step | 0.0209 | 0.213 | 0.098 | 0.922 |
| Hand Height | 0.217 | 0.229 | 0.946 | 0.344 |
| Body Angle | 0.2681 | 0.241 | 1.114 | 0.265 |

By inspecting the p-values in the regression summary, we can see that none of the features are significant predictors. This would suggest that body pose features have no significant effect on penalty save probability. This is also mentioned in [24] which discusses how penalties are more based on game theory instead of the technique of the goalkeeper. However, some of the coefficient signs do align with goalkeeping intuition. Increasing body and torso angle show some evidence of improving save probability and this is consistent with what was discovered using the previous unsupervised learning method in Section 5.2.

### 6.6. Learning From the Pros

In this section, we investigate how our models can be utilised by grassroots players. In this experiment, we collected 18 images from footage of an amateur goalkeeper making 1v1 saves in a university training session. Our models can be used to analyse goalkeeping technique in both training sessions and matches. Footage needs to be recorded of the goalkeeper making 1v1 saves. Then, using the footage, we can take the image of the goalkeeper when the striker shot the ball. We can also derive basic locational data of the goalkeeper and striker, as well as noting whether the striker was under pressure or not from the footage. Images collected from this process are then fed into the 3D pose estimation model described in Section 4.1 and adapted for view-invariance using methodology in Section 4.2. GKEM is calculated using the recorded locational data. The data collection process is summarised in Figure 13.

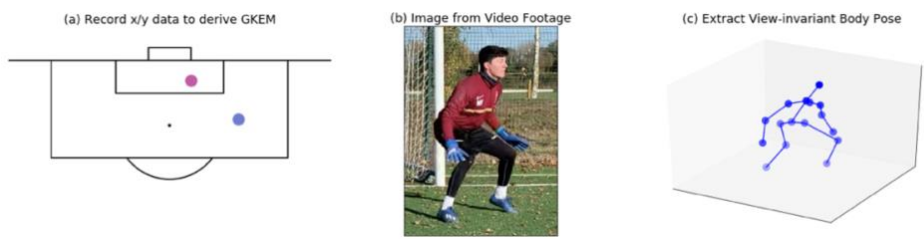

**Figure 13**: Data collection process to enable amateur goalkeepers to utilise our models.

An amateur goalkeeper can use this methodology and these models in a variety of ways. These models highlight 1v1s in which the goalkeeper made a suboptimal decision, allowing the goalkeeper to learn from these mistakes. Our models also suggest which technique would have led to the highest probability of making a save, giving information to the goalkeeper that can help improve their decision-making ability. Figure 14 shows examples of both optimal technique and suboptimal technique used by an amateur goalkeeper.

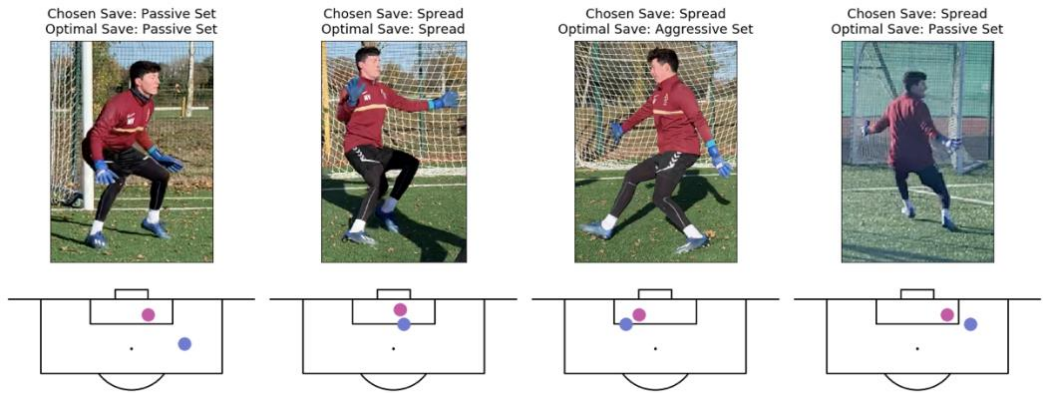

**Figure 14**: Examples of both optimal and sub-optimal technique usage by an amateur goalkeeper.

## 7. Summary

This paper focuses on 1v1 and penalty saving technique, but future work can seek to expand on this by studying save technique from all types of shots, cross collection technique, and distribution technique. With a larger dataset, and a bespoke goalkeeper 3D pose estimation model, the methods introduced in this paper can act as a framework for future analysis in this area. Our methods can also be extended to analysing professional technique in golf swings and baseball pitches, among other sports.

Using a combination of broadcast footage and event data, we have provided a framework for understanding goalkeeper technique and decision-making. By training our models on data collected from Premier League and International goalkeepers, we provide tools for amateur goalkeepers from which to learn. Optimal technique maps can be used for coaching purposes and xSAA assigns a value to save technique choice.

We showed how unsupervised machine learning successfully identifies four 1v1 save techniques using only broadcast footage and shot location data. Trained on 311 shots, we trained an SVM xS model, achieving an accuracy of 68.5% on the test set, inference from this model reveals patterns in optimal save technique. We also found two clusters of penalty save techniques, corresponding to shallow and large diving angles. We found that penalty saves adopting a larger diving angle at the point of impact of the shot had a larger save percentage by 3.6%.

We have presented a variety of applications for how these methods can be applied in practice. It enables amateur goalkeepers and professional goalkeepers alike to use quantitative analysis to study and improve their game. The models can also be used as a new toolkit for scouting goalkeepers and assisting analysts to pinpoint match footage in which a goalkeeper employs sub-optimal technique choice. They allow an analyst to evaluate and compare the ability of goalkeepers to make the correct decision in the technique they choose.

## References

[1] Ryan Beal, Timothy J. Norman, and Sarvapali D. Ramchurn. Artificial intelligence for team sports: a survey. The Knowledge Engineering Review, 34:e28, 2019.
[2] Tom Decroos, Lotte Bransen, Jan Van Haaren, and Jesse Davis. Actions speak louder than goals. Proceedings of the 25th ACM SIGKDD International Conference on Knowledge Discovery & Data Mining, 2019.
[3] Ryan Beal, Narayan Changder, Timothy Norman, and Sarvapali Ramchurn. Learning the value of teamwork to form efficient teams. Proceedings of the AAAI Conference on Artificial Intelligence, 34(05):7063–7070, 2020.
[4] Beal, R., G. Chalkiadakis, T.J. Norman, & S.D. Ramchurn, Optimising Game Tactics for Football, Proceedings of the 19th International Conference on Autonomous Agents and MultiAgent Systems, 2020.
[5] Javier Fernández and Luke Bornn. Wide open spaces: A statistical technique for measuring space creation in professional soccer, 2018.
[6] P. Power, A. Cherukumudi, Sujoy Ganguly, Felix Wei, Long Sha, Jennifer Hobbs, Héctor Ruiz, and P. Lucey. Trading places – simulating goalkeeper performance using spatial & body-pose data. 2019.
[7] Panna Felsen. "body shots": Analyzing shooting styles in the nba using body pose. 2017.
[8] Adrià Arbués-Sangües, Gloria Haro, Coloma Ballester, and Adrián Martín. Head, shoulders, hip and ball... hip and ball! using pose data to leverage football player orientation, 2019.
[9] Adrià Arbués-Sangües, Adrián Martín, Paulino Granero, Coloma Ballester, and Gloria Haro. Learning football body-orientation as a matter of classification, 2021.
[10] D. Yam. A data driven goalkeeper evaluation framework. 2019.
[11] A. Rathke. An examination of expected goals and shot efficiency in soccer. *Journal of Human Sport and Exercise*, 12:514–529, 2017.
[12] Leonardo Lamas, Rene Drezner, Guilherme Otrando, and Junior Barrera. Analytic method for evaluating players' decisions in team sports: Applications to the soccer goalkeeper. *PLOS ONE*, 2018.
[13] Xingyi Zhou, Qixing Huang, Xiao Sun, Xiangyang Xue, and Yichen Wei. Towards 3d human pose estimation in the wild: a weakly-supervised approach, 2017.
[14] Jun Liu, Henghui Ding, Amir Shahroudy, Ling-Yu Duan, Xudong Jiang, Gang Wang, and Alex C. Kot. Feature boosting network for 3d pose estimation, 2019.
[15] Julieta Martinez, Rayat Hossain, Javier Romero, and James J. Little. A simple yet effective baseline for 3d human pose estimation, 2017.
[16] Aiden Nibali, Zhen He, Stuart Morgan, and Luke Prendergast. 3d human pose estimation with 2d marginal heatmaps, 2018.
[17] Denis Tome, Chris Russell, and Lourdes Agapito. Lifting from the deep: Convolutional 3d pose estimation from a single image. *2017 IEEE Conference on Computer Vision and Pattern Recognition (CVPR)*, 2017.
[18] S. Lloyd. Least squares quantization in pcm. *IEEE Transactions on Information Theory*, 28(2):129–137, 1982.
[19] David Arthur and Sergei Vassilvitskii. K-means++: The advantages of careful seeding. In Proceedings of the Eighteenth Annual ACM-SIAM Symposium on Discrete Algorithms, SODA '07, page 1027–1035, USA, 2007.

[20] van der Maaten, L.J.P.; Hinton, G.E. Visualizing High-Dimensional Data Using t-SNE. Journal of Machine Learning Research 9:2579-2605, 2008.
[21] Peter J. Rousseeuw. Silhouettes: A graphical aid to the interpretation and validation of cluster analysis. Journal of Computational and Applied Mathematics, 20:53–65, 1987.
[22] Nello Cristianini and Elisa Ricci. Support Vector Machines, pages 928–932. Springer US, Boston, MA, 2008.
[23] John Platt. Probabilistic outputs for support vector machines and comparisons to regularized likelihood methods. Adv. Large Margin Classif., 10, 2000.
[24] Karl Tuyls, Shayegan Omidshafiei, Paul Muller, Zhe Wang, Jerome Connor, Daniel Hennes, Ian Graham, William Spearman, Tim Waskett, Dafydd Steele, Pauline Luc. Game plan: What ai can do for football, and what football can do for ai, 2020.

# Appendix 1 – Body Pose Keypoints

The PoseHG3D predicts the 3D coordinates of 16 body pose keypoints, illustrated in Figure 15. Here, we define a notation for the coordinates of the individual body keypoints that is used to define the penalty save feature space in Appendix 2. Let $p_j = [x_j, y_j, z_j]$, where $j \in joints$ is the set of 16 joints.

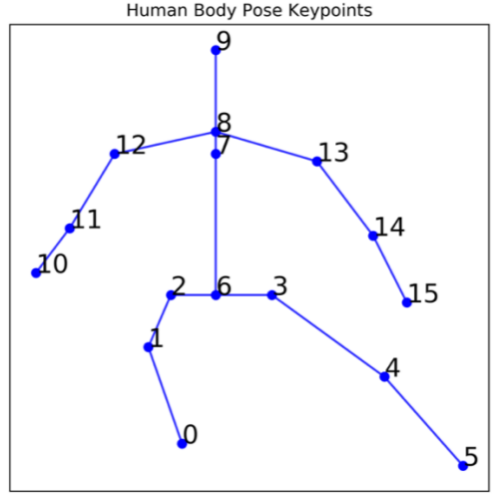

**Figure 15**: The 3D pose estimation model predicts 16 body pose keypoints, labelled in this illustration. Here, $p_{10}$ is the right hand of the goalkeeper.

# Appendix 2 – Penalty Save Feature Space

Descriptions of the five hand-crafted features that make up the penalty save feature space are given in Table 3.

**Table 3**: Descriptions and equations for the penalty save feature space

| Feature | Equation | Description |
|---|---|---|
| Torso Angle | $\left\lvert \arctan\left(\frac{x_7}{y_7}\right) \right\rvert$ | Angle that the torso makes with the horizontal |
| Body Angle | $\left\lvert \arctan\left(\frac{x_7 - x_{0/5}}{y_7 - y_{0/5}}\right) \right\rvert$ | Angle that the entire body makes with the horizontal |
| Height | $\lvert \max(y_0, y_1, \ldots, y_{16}) - \min(y_0, y_1, \ldots, y_{16}) \rvert$ | Distance of the highest body keypoint to the ground |
| Forward Step | $\lvert z_0 - z_5 \rvert$ | Absolute difference in depth between the goalkeeper's feet |
| Hand Height | $\lvert \min(y_0, y_1, \ldots, y_{16}) - \min(y_{10}, y_{15}) \rvert$ | Distance of the lowest hand to the ground |